\documentclass[11pt,a4paper]{article}
\usepackage[hyperref]{eacl2021}
\usepackage{times}
\usepackage{latexsym}

\usepackage{arydshln}
\usepackage{booktabs}
\usepackage{graphicx}

\usepackage{microtype}

\aclfinalcopy % Uncomment this line for the final submission
\def\aclpaperid{19} %  Enter the acl Paper ID here

\usepackage{booktabs}
\usepackage{xcolor}
 
\title{BembaSpeech: A Speech Recognition Corpus for the Bemba Language}

\author{Claytone Sikasote\thanks{\ \ Work done while at African Masters in Machine Intelligence (AMMI).}\\
  Department of Computer Science \\
  University of Zambia  \\
  Zambia \\
  \texttt{claytone.sikasote@cs.unza.zm} \\\And
  Antonios Anastasopoulos \\
  Department of Computer Science  \\
  George Mason University \\
  USA \\
  \texttt{antonis@gmu.edu} \\}

\date{}

\begin{document}
\maketitle
\begin{abstract}
Ili ipepala lilelanda pamashiwi  mu Cibemba na ifyebo fyalembwa ifyabikwa pamo nga mashiwi yakopwa elyo na yalembwa  ukupanga  iileitwa  BembaSpeech. Iikwete amashiwi ayengabelengwa ukufika kuma awala amakumi  yabili na yane mu lulimi lwa Cibemba, ululandwa na impendwa ya bantu ba mu Zambia ukufika  cipendo  ca 30\%. Pakufwaisha ukumona ubukankala bwakubomfiwa mu mukupanga  ifya mibombele ya ASR  mu Cibemba, tupanga imibombele ya ASR iya  mu Cibemba ukufuma pantendekelo ukufika na pampela,  kubomfya elyo na ukuwaminisha icilangililo ca mibomfeshe yamashiwi na  ifyebo ifyabikwa pamo mu Cisungu icitwa DeepSpeech na ukupangako iciputulwa ca mashiwi na ifyebo fyalembwa mu Cibemba (BembaSpeech corpus). Imibobembele yesu iyakunuma ilangisha icipimo ca kupusa nelyo ukulufyanya  kwa mashiwi ukwa 54.78\%. Ifyakufumamo filangisha ukuti ifyalembwa kuti fyabomfiwa ukupanga imibombele ya ASR mu Cibemba.

We present a preprocessed, ready-to-use automatic speech recognition corpus, BembaSpeech, consisting over 24 hours of read speech in the Bemba language, a written but low-resourced language spoken by over 30\% of the population in Zambia. To assess its usefulness for training and testing ASR systems for Bemba, we train an end-to-end Bemba ASR system by fine-tuning a  pre-trained DeepSpeech English model on the training portion of the BembaSpeech corpus. Our best model achieves a word error rate (WER) of 54.78\%. The results show that the corpus can be used for building ASR systems for Bemba.\footnote{The corpus and models will be publicly released~\url{https://github.com/csikasote/BembaSpeech}.}

\end{abstract}
\section{Introduction}
Speech-to-Text, also known as Automatic Speech Recognition(ASR) or simply just Speech Recognition (SP), is the task of recognising and transcribing spoken utterances into text. In recent years, there has been a tremendous growth in popularity of speech-enabled applications. This can be attributed to their usability and integration across wide domain applications, such as voice over control systems. However, building well-performing ASR systems typically requires massive amounts of transcribed speech, as well as large text corpora. This is generally not an issue for well-resourced languages such as English and Chinese, where ASR applications have been successfully built with remarkable results~\cite[et alia]{Amodei2016}.

Unfortunately, this is not the case for Africa and its over 2000 languages~\cite{Heine2000}. The prevalence of speech recognition applications for African languages is very low. This can at least partially be attributed to the lack or unavailability of linguistic resources (speech and text) for most African languages~\cite{Martinus2019}. This is particularly the case with Zambian languages. There exist no general speech or textual datasets curated for building natural language processing systems, including ASR systems.

In this paper we present a speech corpus, BembaSpeech, consisting of over 24 hours of read speech in Bemba, a written but under-resourced language spoken by over 30\% of the population in Zambia. We also present an end-to-end speech recognition Bemba model obtained by fine-tuning a pre-trained DeepSpeech English model on BembaSpeech corpus. To our knowledge this is the first work carried out towards building ASR systems for any Zambian language.

The rest of the paper is organized as follows. In section 2, we summarise similar works in ASR for under-resourced languages with a focus on Africa languages. In Section 3 we provide details on the Bemba language. In section 4, we outline the development process of the BembaSpeech corpus, and in section 5 we provide details of our experiments towards building a Bemba ASR model. Last, section 6 discusses our experimental results,  before drawing conclusions and sketching out future research directions.

\section{Related Work}
In the recent past, despite the challenge of limited availability of linguistic resources, several works have been carried out to improve the prevalence of ASR applications in Africa. For example, \citet{Gauthier2016} collected speech data and developed ASR systems for four languages: Wolof, Hausa, Swahili and Amharic. In South Africa, researchers~\citep{DeWet1999, Badenhorst2011, Henselmans2013, VanHeerden2016, DeWet2017} have investigated and built speech recognition systems for South African languages.  Other languages that have seen development of linguistic resources for ASR applications include: Fongbe~\cite{Laleye2016} of Benin; Swahili~\cite{Gelas2012} predominantly spoken by people of East Africa; Amharic, Tigrigna, Oromo and Wolaytta of Ethiopia~\citep{Abate2005, Tachbelie2014, Abate2020, Woldemariam2020}; Hausa\cite{Schlippe2012} of Nigeria and Somali~\cite{Abdillahi2006} of Somalia. 
In all the aforementioned works, Hidden Markov Models~\cite{Juang1991} and traditional statistical language models are adopted to develop ASR systems, typically using the Kaldi~\cite{Povey2011} or HTK~\cite{Young2009} frameworks. The disadvantage of such approaches is that they typically require separate training for all their pipeline components including the acoustic model, phonetic dictionary, and language model. 

Recently, end-to-end deep neural network approaches have successfully been applied to speech recognition tasks~\citep[et alia]{Amodei2016, Pratap2018} achieving remarkable results outperforming traditional HMM-GMM approaches. Such methods require only a speech dataset with speech utterances and their transcriptions for training. In this work, we use an open source end-to-end neural network system, Mozilla`s DeepSpeech~\cite{Hannun2014} to develop a Bemba ASR model using our BembaSpeech corpus.

\section{Bemba Language}
The language we focus on is Bemba (also referred to as ChiBemba, Icibemba), a Bantu language principally spoken in Zambia, in the Northern, Copperbelt, and Luapula Provinces. It is also spoken in southern parts of the Democratic Republic of Congo and Tanzania. It is estimated to be spoken by over 30\% of the population of Zambia~\citep{Kula2008,Kapambwe2018}.

Bemba has 5 vowels and 19 consonants~\cite{Spitulnik2001}. Its syllable structure is characteristically open and is of four main types: V, CV, NCV, and NCGV (where V = vowel (long or short), C = consonant, N = nasal, G = glide (w or y))\cite{Spitulnik2014}. The writing system is based on Latin script~\cite{Mwansa2017}.

Similar to other Bantu languages, Bemba is described to have a very elaborate noun class system which involves pluralization patterns, agreement marking, and patterns of pronominal reference. There are 20 different classes in Bemba: 15 basic classes, 2 subclasses, and 3 locative classes~\cite{Spitulnik2001,SpitulnikVidali2014}. Each noun class is indicated by a class prefix (typically VCV-, VC-, or V-) and the co-occurring agreement markers on adjectives, numerals and verbs.

In terms of tone, Bemba is considered to be a tone language, with two basic tones, high (H) and low (L)~\cite{Kula2016}. A high tone is marked with an acute accent (e.g. \'{a}) while a low tone is typically unmarked. As with most other Bantu languages, tone can be phonemic and is an important functional marker in Bemba, signaling semantic distinctions between words~\cite{Spitulnik2001,SpitulnikVidali2014}.

\section{The BembaSpeech Corpus}
\label{sec:length}
\paragraph{Description}
The corpus has a size of 2.8 GigaBytes with a total duration of speech data of approximately over 24 hours. We provide fixed train, development, and test splits to facilitate future experimentation. The subsets have no speaker overlap among them. Table ~\ref{tab:corpus} summarises the characteristics of the corpus and its subsets. All audio files are encoded in Waveform Audio File Format (WAVE) with a single track (mono) and recording with a sample rate of 16kHz.

\begin{table*}
	\centering
	\begin{tabular}{cccccc}
		\toprule
		\toprule
		\textbf{Subset} & \textbf{Duration} & \textbf{Utterances} & \textbf{Speakers} & \textbf{Male} & \textbf{Female} \\
		\midrule
		\multicolumn{6}{l}{Whole Corpus:}\\
		\cmidrule(l){1-2} 
		Train & 20hrs & 11906 & 8 & 5 & 3\\
		Dev & 2hrs, 30min & 1555 & 7 & 3 & 4\\
		Test & 2hrs & 977 & 2 & 1 & 1\\
		\midrule
		Total & 24hrs, 30min & 14438 & 17 & 9 & 8\\
		\midrule
		\multicolumn{6}{l}{Used in our experiments:}\\
		\cmidrule(l){1-2} 
		Train & 14hrs, 20min & 10200 & 8 & 5 & 3\\
		Dev & 2hrs & 1437 & 7 & 3 & 4\\
		Test & 1hr, 18min & 756 & 2 & 1 & 1\\
		\midrule
		Subset total & 17hrs, 38min & 12393 & 17 & 9 & 8\\
		\bottomrule
	\end{tabular}
	\caption{\label{tab:corpus}General characteristics of the BembaSpeech ASR corpus. We use a subset (audio files shorter than 10 seconds) for our baseline experiments.}
\end{table*}

\paragraph{Data collection}
To build the BembaSpeech corpus we used the Lig-Aikuma app~\cite{Gauthier2016} for recording speech. Speakers used the elicitation mode of the software to record audio from text scripts tokenized at sentence level. The Lig-Aikuma has been used by other researchers for similar works~\citep{Blachon2016, Gauthier2016a, Gauthier2016b}.

\paragraph{Speakers}
The speakers involved in BembaSpeech recording were students of Computer Science in the School of Natural Science at the University of Zambia. The corpus consists of 14,438 audio files recorded by 17 speakers, 9 male and 8 female. Based on the information extracted from metadata as supplied by speakers, their range of age is between 22 and 28 years and all of them identified as black. All the speakers were selected based on their fluency to speak and read Bemba and are not necessarily native language speakers. There are 14 native Bemba speakers, 1 Lozi, 1 Lunda and 1 Nsenga. It is also important to note that the recordings in this corpus were conducted outside controlled conditions. Speakers recorded as per their comfort and have varied accents. Therefore, some utterances are expected to have some background noise. We consider this ``more of a feature than a bug" for our corpus: it will allow us to train and, importantly, evaluate ASR systems that match \textit{real-world} conditions, rather than a quiet studio setting.

\paragraph{Preprocessing}
The corpus was preprocessed and validated to ensure data accuracy by eliminating all corrupted audio files and, most importantly, to ensure that all utterances matched the transcripts. All the numbers, dates and times in the text were replaced with their text equivalent according to the utterances. We also sought to follow the LibriSpeech~\cite{Panayotov2015} file organization and nomenclature by grouping all the audio files according to the speaker, using speaker ID number. In addition, we renamed all the audio files by pre-pending the speaker ID number to the utterance ID numbers.  

\paragraph{Text Sources}
The phrases and sentences recorded were extracted from diverse sources in Bemba language, mainly Bemba literature. In Table~\ref{tab:sources}, we summarise the sources of text contained in BembaSpeech. The length of the phrases varies from a single word to as many as 20 words.

\paragraph{Availability}
The corpus is made available to the research community licensed under the Creative Commons BY-NC-ND 4.0 license and it can be found at our github project repository.

\begin{table}[t]
	\centering
	\begin{tabular}{clc}
		\toprule
		\textbf{ID} & \textbf{Source Name} & \textbf{Size(\%)} \\ 
		\midrule
		1 & Bemba literature & 70\\
		2 & Other online resources & 15 \\
		3 & Local Radio/TV shows & 10 \\
		4 & Youtube movie &  5 \\
		\bottomrule
	\end{tabular}
	\caption{\label{tab:sources} Sources of text contained in BembaSpeech corpus. The Bemba literature includes publicly available books, magazines and training materials written in Bemba. Other online resources includes various websites with Bemba content.}
\end{table}

\section{Experiments}
In this section, we describe the experiments to ascertain the usefulness of the speech corpus for ASR applications.\footnote{Code to reproduce our experiments is available here: \url{https://github.com/csikasote/BembaASR}.}
\subsection{DeepSpeech Model}
In our experiments,  we use Mozilla`s DeepSpeech - an open source implementation of a variation of Baidu`s first DeepSpeech paper~\cite{Hannun2014}. This architecture is an end-to-end sequence-to-sequence model trained via stochastic gradient descent~\cite{Bottou2012} with the Connectionist Temporal Classification~\cite[CTC]{Graves2006} loss function. The model is six layers deep:  three fully connected layers connected followed by a unidirectional LSTM~\cite{Hochreiter1997} layer followed by two more fully connected layers. All hidden layers have a dimensionality of 2048 and a clipped ReLU~\cite{Nair2010} activation. The output layer has as many dimensions as characters in the alphabet of the target language (including  desired punctuations and blank symbols used for CTC). The input layer accepts a vector of 19 spliced frames (9 past frames, 1 present frame and 9 future frames) with 26 MFCC features each. We use the DeepSpeech v0.8.2\footnote{\url{https://github.com/mozilla/DeepSpeech/tree/v0.8.2}} release for all our experiments.

\subsection{Data preprocessing}
We preprocessed the data in conformity with the expectation of the DeepSpeech input pipeline. We converted all transcriptions to lower case. Since DeepSpeech only accepts audio files not exceeding 10 seconds, we considered only audio files with that duration for our training. This resized the corpus for training as can be seen in Table~\ref{tab:corpus}. We also generated an alphabet of characters and symbols which appear in the text, the length of which determines the size of the output layer of the DeepSpeech model. We note that, since Bemba uses the Latin alphabet, our alphabet was the same as that of the pretrained DeepSpeech English model.

\iffalse
\begin{table*}[t]
	\centering
	\begin{tabular}{cccccc}
		\toprule
		\textbf{Subset} & \textbf{Size} & \textbf{No. of Audio Files} & \textbf{No. of Speakers} & \textbf{Male} & \textbf{Female} \\
		\midrule
		Train & 14hrs, 20min & 10200 & 8 & 5 & 3\\
		Dev & 2hrs & 1437 & 7 & 3 & 4\\
		Test & 1hr, 18min & 756 & 2 & 1 & 1\\
		\midrule
		Total & 17hrs, 38min & 12, 393 & 17 & 9 & 8\\
		\bottomrule
	\end{tabular}
	\caption{\label{citation-guide}
		The 17hrs, 38min speech dataset on which we used for benchmarking our Bemba ASR system (a subset of the complete BembaSpeech consisting only of audio files shorter than 10 seconds).
	}
\end{table*}
\fi

\begin{table}[t]
	\centering
	\begin{tabular}{cclc}
		\toprule
		\multicolumn{4}{r}{\textbf{No. of Tokens}} \\
		\cmidrule(r){3-4} 
		\textbf{Language Model} & \textbf{Sentences} & \textbf{Unique} & \textbf{Total} \\ 
		\midrule
		LM1 & 13461 & 27K & 123K \\
		LM2 & 403452 & 189K & 5.8M \\
		\bottomrule
	\end{tabular}
	\caption{\label{tab:LMstats}The token counts for the two sets of text sources used to create the language models.}
\end{table}

\subsection{Training a Bemba Model from Scratch}
Similar to~\cite{Hjortnaes2020, Meyer2020}, we trained DeepSpeech from scratch using the default parameters\footnote{With the exception of batch size: instead of using the default batch size of 1 for train, dev and test, we used 64, 32, 32 respectively for all our experiments.} on the BembaSpeech dataset, providing a baseline model for our experiments. 

\subsection{Transfer Learning from English to Bemba}
In our search for a better performing model, we applied and also experimented with cross-lingual transfer learning. We achieve this by fine-tuning a well performing DeepSpeech English pre-trained model on our Bemba dataset, using a learning rate of 0.00005, dropout at 0.4, and 50 training epochs with early stopping.  

\subsection{Auxillary Language Model}
Similar to the original DeepSpeech approach presented by~\citet{Hannun2014}, we considered including the language model to the acoustic model to improve performance. In order to identify the language model that most improved model performance, we evaluated two sets of language models each consisting, 3-gram, 4-gram and 5-gram. The first set of language models, denoted LM1, were generated from text sourced from train and development transcripts. The second set, denoted LM2, were sourced from a combination of text from train and development transcripts and additional Bemba text from the JW300 dataset~\cite{Agic2020}. In Table~\ref{tab:LMstats} we give the token count for LM1 and LM2. All the language models were generated using the KenLM~\cite{Heafield2011} language model library. We used the DeepSpeech native library to create the trie based models with default parameter values. The same speech recognition model obtained from section 5.4 was used changing only the language model.

\begin{table}[t!]
	\centering
	\begin{tabular}{ccc}
		\toprule
		\textbf{Model} & \textbf{WER(\%)} & \textbf{CER\%} \\ 
		\midrule
		BL & 1.00 & 85.67 \\
		FT & 71.21 & 16.68 \\
		FT + LM1-3 & 54.79 & 18.54 \\
		FT + LM1-4 & 54.80 & 18.08 \\
		FT + LM1-5 & \textbf{54.78} & \textbf{17.05} \\
		FT + LM2-3 & 55.65 & 19.69 \\
		FT + LM2-4 & 55.84 & 20.49 \\
		FT + LM3-5 & 55.75 & 19.99 \\
		\bottomrule
	\end{tabular}
	\caption{\label{tab:results} The best results from our experiments were obtained through fine-tuning a pretrained model and combining it with a 5-gram LM generated only from transcripts. In the table, BL denotes a baseline model and FT denotes fine-tuned model.}
\end{table}

\section{Results}
Table~\ref{tab:results} summarises the results obtained from our experiments.  The best performing model was FT + LM1-5, obtained from finetuning DeepSpeech with a 5-gram language model generated from text sourced from transcripts. The model achieved a word error rate (WER) of 54.78\% and character error rate (CER) of 17.05\%. 

The results also show the impact of the language model on improving the performance of the Bemba ASR model. By including the language model we were able to improve the model performance by a significant margin from 71.21\% to 54.78\% WER. Interestingly, no significant change in performance was observed by the inclusion of the additional 389,991 sentences from the JW300 Bemba data.

\section{Conclusion}
In this paper, we presented an ASR corpus for Bemba language, BembaSpeech. We also demonstrated the usefulness of the corpus by building an End-to-End Bemba ASR model obtained by finetuning a well performing DeepSpeeh English model on the 17.5 hours speech dataset, a subset of BembaSpeech.

For the future, there are many things we can do to improve the results of our model. We are interested to tune the models (both acoustic and language) by expanding the parameter search. We are also interested in further improving our corpus both in size and number of speakers involved. In addition, it would be interesting to compare these results with other frameworks like the Facebook`s wav2letter++ \cite{Pratap2018} and Pytorch-Kaldi~\cite{Ravanelli2019}. 

Lastly, we plan to (a) collect even more data in Bemba, (b) collect data in the different Bemba varieties as spoken throughout Zambia, as well as (c) other Zambian languages.

%\an{Perhaps we can also mention that we would like to (a) collect even more data in Bemba, (b) collect data in the dialects of Bemba, as well as (c) other Zambian languages.}

\ifaclfinal
\section*{Acknowledgments}
We would like to express our gratitude to all the speakers who were involved in the creation of our corpus. This work would not have been successful without their time and effort. We also want to thank Eunice Mukonde-Mulenga for her help with the Bemba translation of the abstract. Twatotela saana. Thank you!
\fi

\bibliography{eacl2021}
\bibliographystyle{acl_natbib}
\end{document}